# Probabilistic Video Generation using
# Holistic Attribute Control

Jiawei He[1★], Andreas Lehrmann[2], Joseph Marino[3], Greg Mori[1], Leonid Sigal[4]

Simon Fraser University[1], Facebook[2], California Institute of Technology[3],
The University of British Columbia[4]

**Abstract.** Videos express highly structured spatio-temporal patterns of visual data. A video can be thought of as being governed by two factors: (i) temporally invariant (*e.g.*, person identity), or slowly varying (*e.g.*, activity), attribute-induced appearance, encoding the persistent content of each frame, and (ii) an inter-frame motion or scene dynamics (*e.g.*, encoding evolution of the person executing the action). Based on this intuition, we propose a generative framework for video generation and future prediction. The proposed framework generates a video (short clip) by decoding samples sequentially drawn from a latent space distribution into full video frames. Variational Autoencoders (VAEs) are used as a means of encoding/decoding frames into/from the latent space and RNN as a way to model the dynamics in the latent space. We improve the video generation *consistency* through temporally-conditional sampling and *quality* by structuring the latent space with attribute controls; ensuring that attributes can be both inferred and conditioned on during learning/generation. As a result, given attributes and/or the first frame, our model is able to generate diverse but highly consistent sets of video sequences, accounting for the inherent uncertainty in the prediction task. Experimental results on Chair CAD [1], Weizmann Human Action [2], and MIT Flickr [3] datasets, along with detailed comparison to the state-of-the-art, verify effectiveness of the framework.

## 1 Introduction

Deep generative models, such as variational autoencoders (VAEs) [4] and generative adversarial networks (GANs) [5], have recently received increased attention [6,7,8,9] due to their probabilistic and unsupervised nature and their ability to synthesize large numbers of interdependent variables from compact representations. Impressive results have been achieved in a broad range of domains, including image generation [10], text synthesis [11], and text-based image synthesis [12,13].

Despite the impressive progress towards better image generation, including controlled attribute-based models [13,14], it still remains a challenge to generate videos. Video generation models are inherently useful for building spatio-temporal priors, forecasting [7,9,15], and unsupervised feature learning [16]. Although a video can usually be represented as a sequence of temporally coherent images, the extension from image generation to video generation is surprisingly difficult.

In videos, in addition to individual frames containing plausible object/scene arrangements, the motions of those objects and scene elements, over time, need to be

---

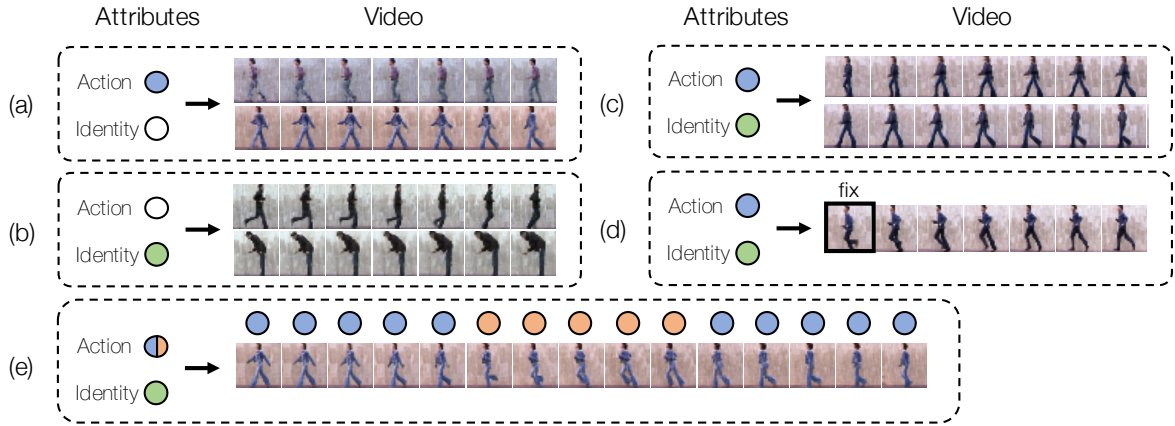

Fig. 1: **Video Generation using Attribute Control.** Our framework uses a semi-supervised latent space containing a fixed number of control signals to steer the generation. Setting one **(a-b)** or both **(c)** of the attributes 'action' and 'identity' to a desired or inferred category (colored circles) constrains the generative process, but takes advantage of the remaining degrees of freedom to synthesize diverse video samples. In **(d)**, conditioning on both attributes as well as the first frame effectively removes all uncertainty (degrees of freedom) in the generation process. In **(e)** attribute transition from 'walking' to 'running' and back to 'walking' is induced at the 6th and 11th frame, resulting in the illustrated corresponding transition within the generated video.

coherent and plausible as well. This is complicated because some motions might be very local (smile on a face), while others global (waves running onto a beach). Further, there are inherent ambiguities in the potential resulting motion patterns. Meaning, given the same input (*e.g.*, first frame of a person standing) a multitude of plausible futures may realistically unfold (*e.g.*, he/she may continue to stand, may start walking, may walk and then sit). Nevertheless, each one of those future predictions is self consistent. For example, once we start predicting that a person is walking, he/she should continue to walk for some nominal number of frames before a transition is plausible. Therefore a generative video model should have the following properties: (1) it should be able to model diversity of future predictions; (2) each future prediction, which corresponds to a sample from the generative model, should be self-consistent.

We introduce a novel framework **VideoVAE** based on variational autoencoders (VAEs). At each time-step, the VAE encodes the visual input into a high-dimensional latent distribution. This distribution is passed to a long short-term memory (LSTM) to encode the motion expressed in the latent space. At every time-step the resulting latent distribution can be sampled and decoded back into a full image. In order to improve the consistency within a generated sequence and also to control the generation process, we expand the latent space in VAEs into a structured latent space with holistic attribute control. The holistic attribute control can be specified or inferred from data; it can be fixed over time, or can exhibit sparse transitions (see Fig. 1). The hierarchical conditional posterior distributions proposed in the structured latent space thus make predictions conditioned on multiple crucial information sources. In addition, conditional sampling is proposed to utilize the previous samples to generate temporally-coherent sequences. Experiments on three challenging data sets show that these techniques effectively ad-dress criteria (1) and (2) above and can generate promising videos of plausible objects with various motions.

## 2   Related Work

We build upon research in style-content models, deep generative models, semantic latent representations, and video synthesis.

**Style-Content Models.** Our approach is implicitly related to the rich literature on style-content separation (a problem introduced in [17]); in our case a distribution over the *content*, in each frame, is being parameterized by attribute factors that affect the latent state and *style* is modelled by the motion patterns that result from dynamics encoded by an RNN. Bilinear [17], nonlinear [18,19] and factored models [20] have been used in the past, but assumed deterministic linear dynamics in the latent space (*e.g.*, GPDM [19]) and relatively simple temporal signals (*e.g.*, motion capture sequences [19,20] or foreground segmentations [18]).

**Deep Generative Models.** Deep generative models (DGMs) use unlabeled data to learn parameters of a deep topology with compact features. As a prominent member, variational autoencoders (VAEs) optimize the well-known encoder-decoder architecture [4,21] using a variational objective, possibly conditioned on an auxiliary input [22]. Their principled design, as well as generative capabilities, have led to a fast adoption and impressive extensions along several axes: A semi-supervised VAE was proposed in [23]. Hierarchical versions aim at increasing the capacity of VAEs and include [11] and [10]. The expressiveness of the approximate posterior can be increased through normalizing flows [24] and its derivatives [25]. Recurrent frameworks using VAEs as a base model [26] are inherently close to our work. However, previous works do not model the non-trivial nature of videos: objects/scenes remain the same within a short video clip. Also, prior methods usually only aim to model synthesized objects with simple motions.

**Semantic Latent Representations.** Semantic latent spaces are interesting, useful, and have a long history in vision and graphics [27]. Recently, [28] utilized the *graphics code* (a set of pre-defined latent codes) for interpretable representation learning. However, the predefined graphics code constrains the system to the parameterized renderable class of objects. Alternatively, [6] uses mutual information to enforce correspondence between parts of the latent space and attributes. In an unsupervised effort, [29] up-weights the KL divergence term in the variational lower bound; when combined with a standard factorized Gaussian prior, this encourages additional independence between the latent variables. Similarly, [8] uses a hierarchy of latent variables to learn a set of independent hierarchical features. Finally, in [13] a disentangled latent representation is used to generate images conditioned on attributes.

**Video Synthesis.** Several very recent works have been proposed to tackle video synthesis. For example, [9,15] predict uncertain future frames from a static image input. However, such (extremely) short-term predictions are unable to model motion. In a somewhat different task, [30] uses GANs to model motions, and [31] uses VAE to predict trajectories of pedestrians. Related, [32] uses RNNs in the encoder and decoder, as well as a feedforward network in the prior, to model video and other dynamic data, particularly for counter-factual reasoning. In [33], the authors use a VAE to encode linear dynamics in the latent space for videos of basic physics phenomena; [34] uses an additional set of discrete latent variables to model linear dynamics in the latent space. Finally, [35] proposes a probabilistic video model that estimates the discrete joint distribution of the raw pixel values in a video. However, these models lack a natural latent structure to capture semantic-level information.

**Contributions.** Our key contribution is a novel generative video model – **VideoVAE**. VideoVAE is based on the variational autoencoder (VAE), which provides probabilistic methods of encoding/decoding full frames into/from a compact latent state. The motion of the resulting distribution in the latent space, accounting for the motion in the video, is modeled using an LSTM. At every time-step the structured latent distribution can be sampled and decoded back into a full frame. To improve the quality of the inference and generation, we propose a factoring of the latent space into holistic attribute controls and residual information; control variables can either be observed (specified) or inferred from the first frame or snippet of the video (allowing semi-supervised training). Further, since both dynamics and appearance can be multimodal, to avoid jumps among the modes, we propose conditional sampling which facilitates self-consistent sequence generation. Experiments on three challenging datasets show that our proposed model can generate plausible videos that are quantifiably better than state-of-the-art.

## 3    Probabilistic Video Generation

We will now describe our proposed model (Fig. 2). At a high level, VideoVAE models spatio-temporal sequences by building upon VAE as a spatial model and LSTM as a temporal model, *i.e.*, each frame is encoded into a latent distribution (representing appearance dependencies within a frame) that is fed into a recurrent neural network (modeling motion dynamics across frames). We will first provide a brief summary of the two base models (Sec. 3.1) and then discuss our contributions that result in coherent and controllable video generation: a structured latent space with holistic attribute control (Sec. 3.2) and a conditional variational posterior (Sec. 3.3).

### 3.1    Background: Base Models

**Variational Autoencoder (VAE).** A VAE [4] describes an instance of a generative process with simple prior $p_\theta(z)$ (*e.g.*, Gaussian) and complex likelihood $p_\theta(x|z)$ (*e.g.*, a neural network) in which $z$ is a latent and $x$ is an observed variable. Approximating the intractable posterior $p_\theta(z|x)$ with a variational neural network $q_\phi(z|x)$, we can jointly optimize over $\theta$ and $\phi$ by maximizing the variational lower bound $\mathcal{L}$ on the marginal likelihood $p_\theta(x^{(t)})$ of a video frame $x^{(t)}$,

$$\begin{aligned}
\log p_\theta(x^{(t)}) &= \mathrm{KL}(q_\phi \| p_\theta) + \mathcal{L}(\theta, \phi) \\
&\geq \mathcal{L}(\theta, \phi) = -\mathbb{E}_{q_\phi}\left[\log \frac{q_\phi(z|x^{(t)})}{p_\theta(z, x^{(t)})}\right].
\end{aligned} \tag{1}$$

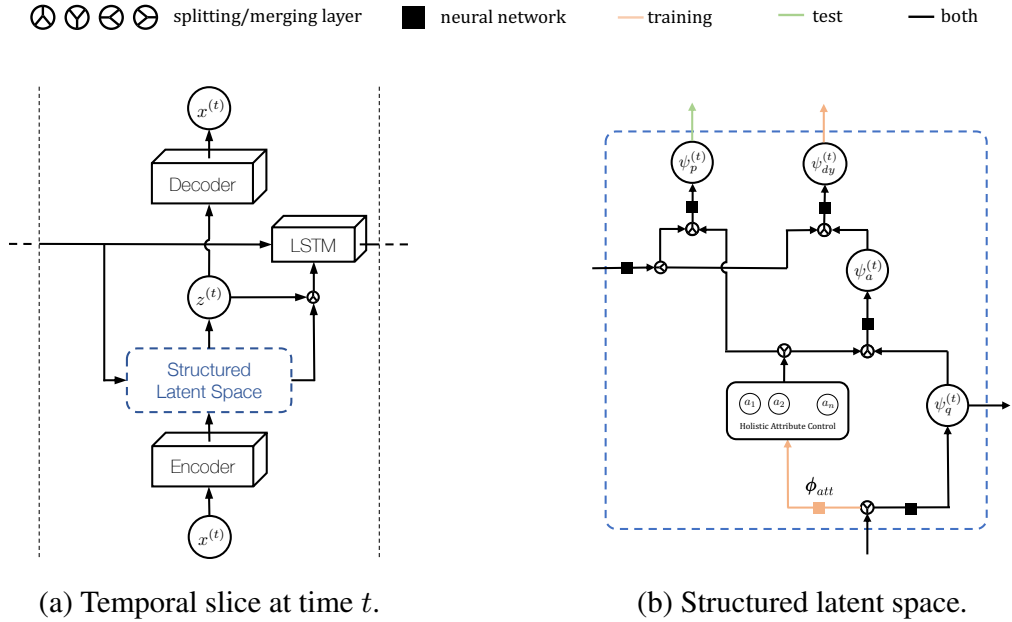

(a) Temporal slice at time $t$.    (b) Structured latent space.

Fig. 2: **Overview. (a)** A structured latent representation of a variational autoencoder (VAE; [4]) encodes a conditional approximate posterior that is propagated through time with the help of a long short-term memory (LSTM; [36]). **(b)** Detailed view of the dashed box in Fig. 2a: In a hierarchical process, holistic attributes are first merged with the variational approximate posterior and then integrated with temporal information from the LSTM, effectively resulting in a doubly-conditional dynamic approximate posterior. We denote the parameters of these distributions as $\psi_{\bullet}^{(t)} := [\mu_{\bullet}^{(t)}, \sigma_{\bullet}^{(t)}]$, where $\mu_{\bullet}^{(t)}$ and $\sigma_{\bullet}^{(t)}$ are the mean and covariance of a multivariate Gaussian, respectively. Information flow only available during specific phases is highlighted as ↑ for training and ↑ for testing. The prior distribution $\psi_{p}^{(t)}$, for instance, is only used to calculate the KL-loss at training time but as a sampling distribution for $z^{(t)}$ at test time.

From an autoencoder point of view, we can think of the approximate posterior $q_{\phi}$ as an encoder and the likelihood $p_{\theta}$ as a decoder. Generating a video frame corresponds to decoding a sample from the prior.

**Long Short-Term Memory (LSTM).** While VAEs are a powerful framework for modeling static video frames, they fail to model the motion dynamics between frames in a video. Long short-term memory (LSTM) [36], a form of a recurrent neural network, is able to capture such dynamic dependencies. An LSTM consists of two components: (1) a transition function $f_h$ that determines the evolution of an internal hidden state; (2) a mapping from the internal hidden state to an output. The transition function of a standard LSTM is entirely deterministic,

$$h^{(t)} = f_h(v^{(t)}, h^{(t-1)}), \qquad (2)$$

where $v^{(t)}$ and $h^{(t)}$ are the LSTM input and hidden state at time $t$, respectively.

### 3.2    Spatial Model

Frames in a video typically exhibit both transient and persistent characteristics. For example, identity and action of a subject are likely to remain fixed (persistent) in a short clip, while the limbs of the person are likely to move (transient) as he/she performs the action. Modeling video using a simple VAE+RNN [26] combination effectively models all frame appearance at the temporal granularity of a frame. This often leads to artifacts during generation, like undesired identity changes. To address this, we structure the latent space by introducing *holistic attribute controls*. The key benefit of such control variables is that they are persistent, meaning that they either stay fixed or change extremely infrequently with respect to the frame rate of the video. The following two paragraphs describe holistic attribute controls in more detail and show their hierarchical integration with residual and temporal information (Fig. 2b).

**Holistic Attribute Control.** Holistic attributes $\mathbf{a} = (a_i)_i$ are a set of predefined attributes that do not change with time.[1] Examples include the person identities in human action sequences or the scene labels in generic video clips. These fixed attribute variables $\mathbf{a}$ cast holistic control on the entire generated video sequence and can, in general, be of various types: categorical, discrete or continuous. Their state can be clamped to a desired value, inferred from data, or even derived from some external data source. In this work, the controls are inferred in a semi-supervised manner at training time and set as fixed during generation.

*Training.* Since the VAE encoder $\phi_{enc}$ already maps the input images $x^{(1:T)}$ to a set of latent features $\phi_{enc}\left(x^{(1:T)}\right)$, we infer the attributes $a_i$ from those representations by adding a small classification network $\phi_{att}^{(i)}$ for each attribute after the encoder.[2] This is illustrated by the lower orange arrow in Fig. 2b and can be expressed as

$$a_i = \phi_{att}^{(i)}\left(\phi_{enc}\left(x^{(1:T)}\right)\right). \tag{3}$$

The image encoder $\phi_{enc}$ and attribute classifiers $\boldsymbol{\phi}_{att} = \{\phi_{att}^{(i)}\}_i$ are learned independently, which allows easy pretraining of the attribute inference and quick adaption to new attributes. Another advantage of this setting is that it makes it possible to utilize a subset of labeled training data to learn $\boldsymbol{\phi}_{att}$ and generalize to the remaining (unlabeled) training instances with the same attributes, leading to a semi-supervised training scenario. In general, we observed that label information for about 20% of the training data is sufficient to infer the remaining attributes. Once the attributes are inferred for each video in the training set, they are used as fixed controls during VideoVAE training (Sec. 4).

*Testing.* The attributes are set as fixed to cast holistic control over the generation process. They could be a single label (*e.g.*, "walking") or a sequence of labels (*e.g.*, "walking" – "running" – "skipping") associated with specific frames to cast transient control.

---

[1] They do not change with time unless explicitly asked to, *e.g.*, to control the temporal content of a synthesized clip, as demonstrated in Fig. 4.

[2] Each network consists of two fully-connected layers with a central ReLU unit, connected in time by an LSTM. This LSTM for attribute inference is independent of the main LSTM modelling motion dynamics.

**Conditional Approximate Posterior.** Traditional VAEs encode the data into an approximate posterior distribution and sample from a prior to synthesize novel data. This works well in image generation, since each synthesized image can be sampled independently. However, in video generation, successive samples should be temporally coherent. In other words, samples should be drawn conditioned on previous information and also the order of the samples matters. The latent code $z$ should combine this type of frame-level consistency with the sequence-level consistency provided by the holistic control variables discussed above.

Based on these observations, we propose the following structured latent space, which comprises a set of hierarchical approximate posterior distributions (Fig. 2b):

(1) an initial approximate posterior distribution, $\mathcal{N}(\mu_q^{(t)}, \sigma_q^{(t)})$, conceptually modeling residual information not captured by holistic attributes;

(2) a conditional approximate posterior, $\mathcal{N}(\mu_a^{(t)}, \sigma_a^{(t)})$, encoding the full appearance of the frame, combining holistic attribute control with the residual posterior above;

(3) a dynamic approximate posterior, $\mathcal{N}(\mu_{dy}^{(t)}, \sigma_{dy}^{(t)})$, which further incorporates motion information and enforces a temporally coherent trajectory. Please refer to Sec. 3.3 for more details on the integration of temporal information.

The three distributions can be expressed in terms of the encoded input, the attributes, and the LSTM state,

$$
\begin{aligned}
\psi_q^{(t)} &= [\mu_q^{(t)}, \sigma_q^{(t)}] = \phi_\tau(\phi_{enc}(x^{(t)})), \\
\psi_a^{(t)} &= [\mu_a^{(t)}, \sigma_a^{(t)}] = \phi_\tau(\psi_q^{(t)}, \mathbf{a}), \\
\psi_{dy}^{(t)} &= [\mu_{dy}^{(t)}, \sigma_{dy}^{(t)}] = \phi_\tau(\psi_a^{(t)}, \phi_\tau(h^{(t-1)})).
\end{aligned}
\tag{4}
$$

Here, $\phi_\tau$ refers to a neural network with an architecture similar to the attribute inference network (two fully-connected layers with a central ReLU unit, but no LSTM). Separate instances of $\phi_\tau$ (black boxes in Fig. 2b) along the hierarchical chain of our structured latent space share this architecture but have different weights.

### 3.3 Temporal Model

VideoVAE contains a VAE at each timestep and propagates information between timesteps with an LSTM to capture the motion dynamics in videos. The following two paragraphs discuss the integration of this temporal information with respect to the encoder and decoder of the VAE at time $t$. An illustration of this interaction is depicted in Fig. 2a.

**Decoder.** The latent variational representations at time-step $t$ are conditioned on the state variable $h^{(t-1)}$ of the LSTM. This additional dependency takes advantage of the fact that videos are highly temporally consistent and prevents the content and motion between two consecutive frames from changing too quickly. As the prior distribution $\psi_p^{(t)}$ represents the model's prediction and belief at timestep $t$ given all previous information, it should not be a fixed Gaussian (as is the case in static VAEs) but follow the distribution

$$
[\mu_p^{(t)}, \sigma_p^{(t)}] = \phi_\tau(\phi_\tau(h^{(t-1)}), \mathbf{a}),
\tag{5}
$$

where $\mu_p^{(t)}$ and $\sigma_p^{(t)}$ denote the parameters of the prior distribution at timestep $t$. With this setting, under the assumption that the LSTM hidden state $h^{(t-1)}$ contains all necessary information from $x^{(<t)}$, the prior distribution at timestep $t$ becomes $p(z^{(t)}|x^{(<t)})$. This distribution changes with time and effectively represents the *prediction* of the current time-step given previous information.

Similarly, the output distribution is updated according to

$$z^{(t)} \sim \mathcal{N}(\mu_{dy}^{(t)}, \sigma_{dy}^{(t)}),$$
$$[\mu_x^{(t)}, \sigma_x^{(t)}] = \phi_{dec}(z^{(t)}), \qquad (6)$$
$$\tilde{x}^{(t)}|z^{(t)} \sim \mathcal{N}(\mu_x^{(t)}, \sigma_x^{(t)}),$$

where $\mu_x^{(t)}, \sigma_x^{(t)}$ denote the parameters of the output distribution and $\tilde{x}^{(t)}$ the reconstruction of the input $x^{(t)}$ at time $t$.

**Encoder.** At each timestep, the frame input $x^{(t)}$ is mapped by the encoder function $\phi_{enc}$ to the hierarchical latent space (Fig. 2b), from which the decoder samples

$$z^{(t)}|x^{(t)} \sim \mathcal{N}(\mu_{dy}^{(t)}, \sigma_{dy}^{(t)}), \qquad (7)$$

where the dynamic approximate posterior is given by Eq. (4).

**Conditional Sampling.** In the temporal framework described thus far, only the distribution $\mathcal{N}(\mu_q^{(t)}, \sigma_q^{(t)})$ is passed to the LSTM. In other words, the samples at each time-step are not passed along time, and are thus independent, resulting in temporally inconsistent sequences (*e.g.*, in terms of the attributes expressed in a decoded RGB frame). Following intuition from LSTM-based language decoders [37], we introduce conditional sampling to address this problem. In addition to the initial approximate posterior distribution $\mathcal{N}(\mu_q^{(t)}, \sigma_q^{(t)})$, sample $z^{(t)}$ is also passed to the LSTM (Fig. 2a). The hidden state of the LSTM is therefore updated according to

$$h^{(t)} = f_h(\psi_q^{(t)}, z^{(t)}, h^{(t-1)}). \qquad (8)$$

In this way, based on the past information, a reasonable initial guess of where the sample $z^{(t+1)}$ should be in $\mathcal{N}(\mu_{dy}^{(t+1)}, \sigma_{dy}^{(t+1)})$ is provided. Notably, due to the VAE structure of our model, this change requires no changes to the architecture itself, as compared to, for example, language translation models. The effectiveness of this conditional sampling scheme to improve the consistency of the generated sequences will be shown in Sec. 5.

## 4    Learning and Synthesis

### 4.1    Learning

We follow a two-stage training strategy. As discussed in Sec. 3.2, we first use approximately 20% of the training data to train the holistic attribute classifiers $\phi_{att}$ using a cross entropy loss. Once $\phi_{att}$ is trained, it is used to infer the holistic attributes for the rest of the training set. This part of the training process is thus semi-supervised. The inferred attributes are then considered as fixed during training of the VideoVAE model, the objective function of which becomes a timestep-wise variational lower bound,

$$\mathcal{L} = \mathbb{E}_{q(z^{(\leq t)}|x^{(\leq t)})}[\sum_{t=1}^{T}(\log p(x^{(t)}|z^{(\leq t)}, x^{(<t)}) \tag{9}$$
$$-\mathrm{KL}(q(z^{(t)}|x^{(\leq t)}, z^{(<t)}))||p(z^{(t)}|x^{(<t)}, z^{(<t)})].$$

We optimize Eq. (9) using a pixel-wise $L_1$-loss with standard SGD techniques [38]. The first part is the log-likelihood of the generated data distribution and the second part is the KL divergence between the prior distribution and the approximate posterior distribution at time-step $t$. A full derivation of the objective function is included in the supplementary material.

### 4.2    Synthesis

In order to generate a video, *i.e.*, at test time, we adapt the architecture described in Sec. 3 as illustrated in Fig. 2: First, the holistic attributes are not inferred from the input data, as is the case during training, because there is no input data at test time. Instead, we choose and fix a set of desired holistic attributes to steer the generative process. Second, the sample is not drawn from the dynamic approximate posterior distribution, but from the prior distribution at each time-step, following standard VAE practice. The samples are then decoded into an output frame and fed back into the network.

We further propose two different methods to initialize the generative process:

–  **Holistic attribute controls only:** in this setting, only partial or full holistic attribute controls are provided (Fig. 1(a)–(c)). The initial LSTM state $h^{(0)}$ is randomly initialized and the first generated frame is sampled and decoded from the distribution $[\mu_p^{(1)}, \sigma_p^{(1)}] = \phi_\tau(\phi_\tau(h^{(0)}), \mathbf{a})$.

–  **Holistic attribute controls & first frame:** in this setting, in addition to the holistic attribute controls, the first frame is also provided (Fig. 1(d)). The first generated frame, in this case, is the reconstruction of the input, and the rest of the generation follows Eq. 5. Typically, conditioning on the first frame improves the generation quality as it provides the framework more precise information.

## 5    Experiments

We conduct qualitative and quantitative experiments on multiple datasets to evaluate the proposed framework. After a brief description of the datasets (Sec. 5.1), we describe our evaluation metrics (Sec. 5.2) and validate our contributions in an ablation study and a comprehensive comparison to various baseline models (Sec. 5.3).

### 5.1   Datasets

We evaluate our model on three datasets: Chair CAD [1], Weizmann Human Action [2], and YFCC [3] – MIT Flickr [16]. These datasets contains various kinds of motion patterns, such as simple rotation, structured human action, or complicated scene-related motions.

**Chair CAD [1].** The dataset contains 1393 chair-CAD models. We follow [39] and use a subset of 809 chair models in our experiments. Each chair model is rendered from 31 azimuth angles and 2 elevation angles at a fixed distance to the virtual camera. The rendered images are cropped to have a small border and resized to a common size of $64 \times 64 \times 3$ pixels by [39]. In our experiments, we divide the length-31 sequence into 2 length-16 sequences starting from the $16^{th}$ frame. Altogether, there are four video sequences per chair model. We randomly pick three of them for training, and the last is used for testing generation (conditioned on the first frame).

**Weizmann Human Action [2].** This dataset contains 90 videos of 9 people performing 10 actions. We cropped each frame to center on the person and resize the frames to the size of $64 \times 64 \times 3$. In order to perform generation conditioned on first frame, we first split each video into training and test subset. The first $2/3$ frames of the video are treated as training sequences, and the last $1/3$ frames of the video are treated as test sequences. Then we sample 20 mini-clips of length-10 from each training video to form a final training set.

**YFCC [3] – MIT Flickr [16].** The dataset contains 35 million clips and we use a pre-processed subset of this dataset provided by [16], of witch the videos have been stabilized by SIFT+RANSAC and each video clip contains 32 frames. We use two scene categories *beach* and *golf* provided in [16]. Note that these scene categories are filtered by a pre-trained Place-CNN model, so the labels are not as accurate as the labels in other datasets.

### 5.2   Evaluation Metrics

Quantitative evaluation of generative models is an inherently challenging task. A good generative model should synthesize samples that are both realistic and diverse. Recently, the *Inception Score (I-score)* [40] has been proposed as an evaluation measure reflecting both these criteria. For static images $x$, it is defined as

$$I = \exp\left(\mathbb{E}_x\left[\mathrm{KL}\left[\rho(y|x)\|\rho(y)\right]\right]\right),\tag{10}$$

where $\rho(y|x)$ is the conditional label distribution of an inception model [41] pre-trained on ImageNet [42]. The entropy of this first term in the KL-divergence measures the confidence of the classifier and the entropy of the second term $\rho(y)$ measures the diversity of the marginal label distribution over all generated samples.

  However, in the video generation field, the lack of a standard model structure and large datasets makes it hard to come up with a universal classifier. Therefore, we pre-train individual classifiers on each dataset. We also believe that the first term $\rho(y|x)$ in Eq. (10) is more important for video generation tasks, since it measures the quality

| | Bound | Static | $-C$ | | $+C$ | |
|---|---|---|---|---|---|---|
| | | | $-S$ | $+S$ | $-S$ | $+S$ |
| Intra-E $\downarrow$ | 1.98 | 40.33 | 17.64 | 7.79 | 14.81 | **5.50** |
| Inter-E $\uparrow$ | 1.39 | 0.42 | 0.73 | 1.35 | 1.02 | **1.37** |
| I-Score $\uparrow$ | 4.01 | 1.28 | 1.83 | 3.63 | 2.56 | **3.94** |

Table 1: **Ablation Study on Chair CAD [39].** We evaluate our contributions individually and in combination. $+C$ and $+S$ indicate conditional sampling and a structured latent space, respectively. The proposed videoVAE model uses both elements ($+C+S$; last column). Arrows indicate whether lower ($\downarrow$) or higher ($\uparrow$) scores are better.

of a generated sequence. Therefore, in addition to the I-score, we analyze both terms separately as follows:

**Intra Entropy.** Intra-entropy measures the conditional label entropy of a set $\{x_i\}_i$ of generated video sequences. Specifically, we use the pre-trained classifier to obtain a conditional distribution over the attributes $\mathbf{a}$ and compute

$$S_{\text{intra}} = \sum_i \mathrm{H}\left[\rho(\mathbf{a}|x_i)\right]. \tag{11}$$

A smaller value of $S_{\text{intra}}$ means that the pre-trained classifier is more confident to classify the generated videos, which indicates that they are more similar to real videos.

**Inter Entropy.** Inter-Entropy measures the label entropy of a set $\{x_i\}_i$ of generated video sequences. The pre-trained classifier assigns a label $a_j \in \mathbf{a}$ to each sequence, which allows us to compute the entropy of the induced distribution $p(\mathbf{a})$ over the labels,

$$S_{\text{inter}} = \mathrm{H}\left[\rho\left(\mathbf{a}\right)\right]. \tag{12}$$

A larger value of $S_{\text{inter}}$ indicates that the distribution over the label space is more uniform, which implies that the generative model can produce diverse samples.

### 5.3   Video Synthesis

**Ablation Study.** In order to demonstrate the contribution and effectiveness of each component of our model, we conduct an ablation study on Chair CAD.

*Variants.* We distinguish five different variations of our model: The static model uses a standard VAE to generate chair images. This model is trained with individual frames in the Chair-CAD dataset. 15 consecutively generated images are then treated as one video sequence. A standard VAE plus temporal model in the form of an LSTM is referred to as $(-C - S)$ in Table 1. An illustration of such a model can be obtained from Fig. 2a by omitting conditional sampling and replacing the structured latent space with a single approximate posterior. Previous VAE-based temporal generative models [26] are of this type. The $(+C - S)$ model adds conditional sampling at each time-step to the temporal model. Specifically, the sample $z^{(t)}$ at time-step $t$ is concatenated with the parameters $\psi_q^{(t)}$ of the latent distribution; the merged information is then treated as the input of the LSTM. The $(-C + S)$ model replaces the latent space in the temporal model with our proposed structured latent space with holistic attribute control. With structured latent space, this model should show superior consistency in a generated video compared with the previous models. Finally, we obtain our full pipeline when both conditional sampling and a structured latent space are used; this is the model shown in Fig. 2. A detailed setup of the individual layers is given in the supplementary material.

| | Chair CAD [1,39] | | |
|---|---|---|---|
| | Bound | Deep Rot. [39] ● | VideoVAE (ours) ● |
| Intra-E ↓ | 1.98 | 14.68 | **5.50** |
| Inter-E ↓ | 1.39 | **1.34** | 1.37 |
| I-Score ↑ | 4.01 | 3.39 | **3.94** |

| | Weizmann Human Action [2] | | | |
|---|---|---|---|---|
| | Bound | MoCoGAN [7] ○ | VideoVAE (ours) ○ | ● |
| Intra-E ↓ | 0.63 | 23.58 | 9.53 | **9.44** |
| Inter-E ↓ | 4.49 | 2.91 | **4.37** | **4.37** |
| I-Score ↑ | 89.12 | 13.87 | 69.55 | **70.10** |

| | YFCC [3] — MIT Flickr [16] | | | |
|---|---|---|---|---|
| | Bound | VGAN [16] ○ | VideoVAE (ours) ○ | ● |
| Intra-E ↓ | 30.34 | 46.96 | 44.03 | **38.20** |
| Inter-E ↑ | 0.693 | **0.692** | 0.691 | **0.692** |
| I-Score ↑ | 1.87 | 1.58 | 1.62 | **1.81** |

**Table 2: Quantitative Results.** We report Intra-E, Inter-E, and I-Score on three different datasets. All scores are calculated on attribute classifiers pre-trained on each dataset; comparison across datasets is therefore meaningless. The bounds are calculated by utilizing those classifiers on the actual test videos, *i.e.*, they reflect the statistics of real videos. Depending on the baseline protocol, we compute VideoVAE results for de-novo synthesis (○) and/or prediction given the first frame (●). Arrows indicate whether lower (↓) or higher (↑) scores are better.

*Results.* As shown in Table 1, each part of our model plays an important role. The static version cannot capture the motion patterns, making it impossible to generate consistent sequences. The $(-C - S)$ variant has an LSTM as a temporal model and can generate relatively consistent video sequences. The performance measures show much better results. However, the single approximate posterior distribution in the latent space cannot separate between different modes. As a consequence, attributes such as actions or identities may change along time. It also tends to generate similar sequences, since the latent space is not uniformly distributed and one or a few modes may take up the majority of the latent space. The structured latent space in the $(-C + S)$ model introduces additional information (holistic attribute control) to the model, and therefore the consistency within a generated video (represented by *Intra-E*) is improved by a large margin. In addition, the attribute control disentangles the latent space to some degree, which empirically prevents the modes from collapsing. The conditional sampling in the $(+C - S)$ model improves the consistency between consequent frames. Finally, the $(+C + S)$ model achieves the best result by combining the benefits of both structured latent space and conditional sampling. Visualizations of all models are included in the supplementary material.

**Baseline Comparison.** We compare our method to three baseline methods. Since some of the baseline models do not provide their detailed training setup, we only conduct comparison with them on the dataset they reported in the paper. Specifically, we use Deep Rotator [39] as the baseline model for Chair-CAD, MoCoGAN [7] for Weizmann Human Action and VGAN [16] for YFCC-MIT Flickr.

*Baseline Models.* Deep Rotator uses a simple autoencoder and LSTM structure to generate rotating chairs. However, this framework is limited to simple motions and difficult to generalize. MoCoGAN decomposes the noise vector in GAN models into motion noise and identity noise to separate identity from motion. There are two major drawbacks with this approach: (1) it shows severe mode collapsing, *i.e.*, focuses on a few major motions only; (2) the decomposition limits the approach to scenes with a clear person-action foreground. VGAN uses a two-stream process to generate foreground and background content separately, then combines them into the final video. Although this framework generates prominent foreground objects, the motion and appearance of these objects are usually distorted, unrealistic and exaggerated. VGAN also requires training a model on each category independently.

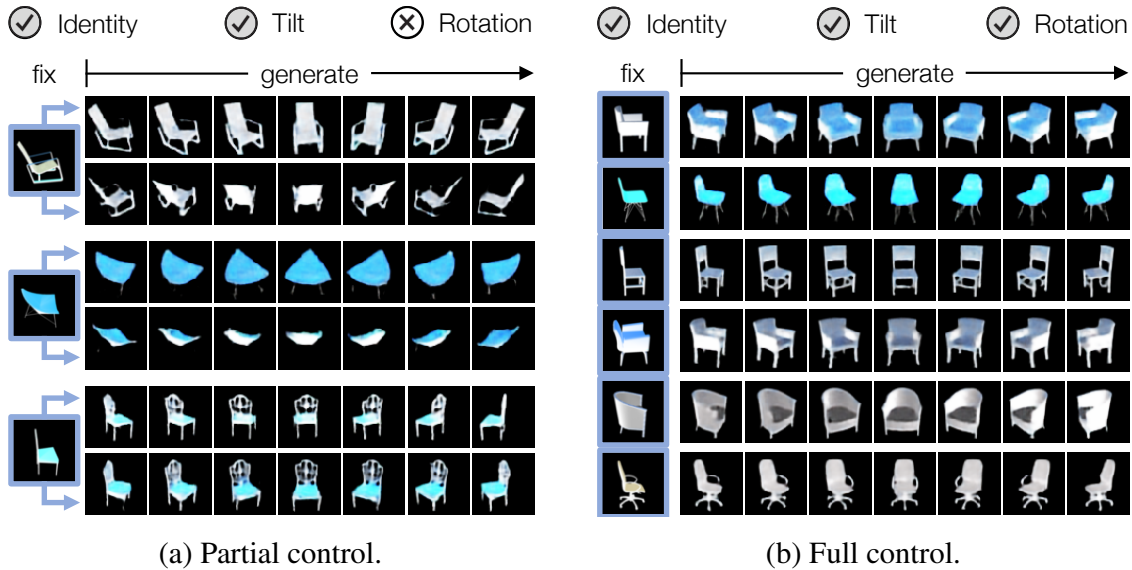

(a) Partial control.                    (b) Full control.

Fig. 3: **Qualitative Results on Chair CAD.**

In our comparison, we use the pre-trained models provided by Deep Rotator and VGAN to generate videos. Since a pre-trained MoCoGAN model is not available, we follow exactly the protocol in [7] to train the model. If the baseline model is deterministic ([16,39]), we generate an equal number of videos for each class for fair comparison. Since the imperfect quality of the generated sequences lowers the classification accuracy, Inter-E and bound are not the same.

*Results.* Our quantitative comparisons with the baseline models are given in Table 2. The proposed VideoVAE model consistently and with a large margin (*e.g.*, inception score: 13.87 (MoCoGAN [7]) vs. 69.55 (ours)) outperforms the baseline models by generating high quality but diverse video sequences. The upper/lower bounds on performance (2nd column in Tables 2/1) are calculated by utilizing the pre-trained attribute classifiers on each dataset's real test set; they represent the statistics of real videos. In addition, Fig. 3 shows the generated sequences on Chair-CAD in various control scenarios: given partial attribute control (chair ID and tilt angle in this case), the model generates chairs rotating to different directions (since direction is unspecified), as shown in Fig. 3a; providing all attribute controls removes the remaining degree of freedom corresponding to direction, resulting in the (unimodal) samples shown in Fig. 3b. Fig. 4 shows the generated sequences on the human action dataset. In Fig. 4a, we fix the action but leave the identity unspecified (partial control), resulting in videos containing differ-ent people performing the same 'jumping jack' action. Also note that holistic controls do not have to be static: Replacing the static controls with a set of time-varying controls (*e.g.*, 'walking-running-walking' instead of 'walking'), we can steer the generative process and synthesize video sequences with smooth transitions between actions, shown in Fig. 4b. Finally, Fig. 5 shows synthetic sequences for two scene types in YFCC-MIT Flickr, illustrating our models capability to handle unconstrained scenes as well. More full-size visualizations and comparisons are provided in the supplementary material.

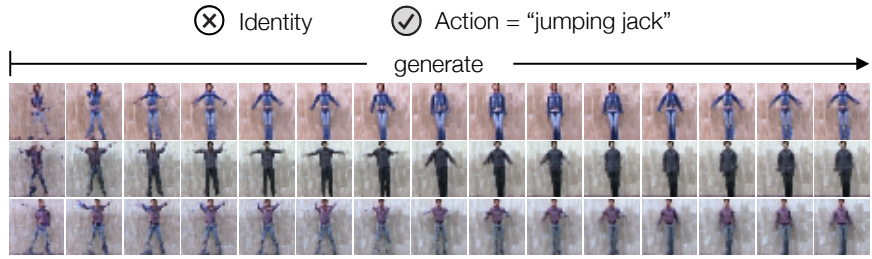

(a) Partial control.

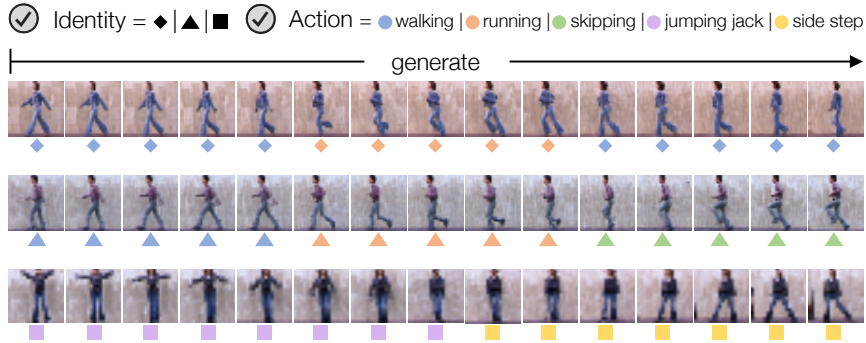

(b) Transient control.

Fig. 4: **Qualitative Results on Weizmann Human Action.** (**a**) Conditioned on one holistic attribute control (action = "jumping jack"), our model generates the corresponding action using different identities. Note that the identity of a person within a sequence is consistent. (**b**) Providing both holistic controls but changing the action attribute during the generation process results in smooth transitions between actions. See supplementary material for full-sized video sequences.

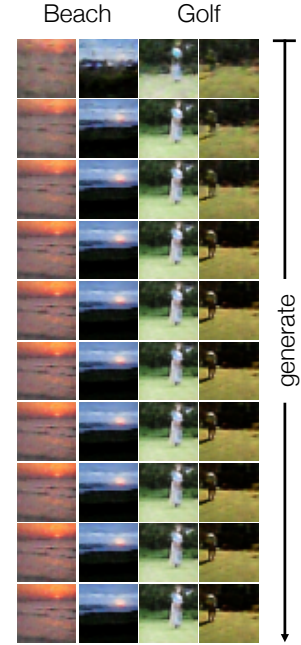

Fig. 5: **Qualitative Results on YFCC – MIT Flickr.** Our model generates realistic clips of both scene types. Each column is one sequence. Refer to the supplementary material for additional visualizations.

*Discussion.* The proposed VideoVAE is an extension of a standard static VAE formulation. Therefore, our model may inherit certain weaknesses of VAEs. While significantly sharper than competing VAE approaches, the output may not always be as crisp as that of latest GANs. Our model also encodes frame motion holistically. In the future, layered variants that separate foreground/background elements, predict their motion and composite them back, should also be explored.

# 6 Conclusion

We propose a novel probabilistic generative framework for video generation and future prediction. The proposed framework generates a video (short clip) by decoding samples sequentially drawn from the latent space distribution into full video frames. VAE is used as a means of encoding/decoding frames into/from the latent space and LSTM as a way to model the distribution dynamics in the latent space. We improve the video generation *consistency* through temporally-conditional sampling and *quality* by structuring the latent space with attribute controls. An ablation study illustrates the importance of our contributions and algorithmic choices. Extensive experiments on three challenging datasets show that our proposed model *significantly* outperforms state-of-the-art approaches in video generation; it also enables controlled generation.

# A    Supplementary Material

We show additional visualizations of the experiments presented in Sec. 5.3 in the main paper and provide additional details about our architecture and objective function.

## A.1    Ablation Study

Visual results corresponding to the four variations of our model ($\pm C \pm S$) are depicted in Fig. 6. As in the main paper, $+C$ and $+S$ refer to conditional sampling and a structured latent space, respectively.

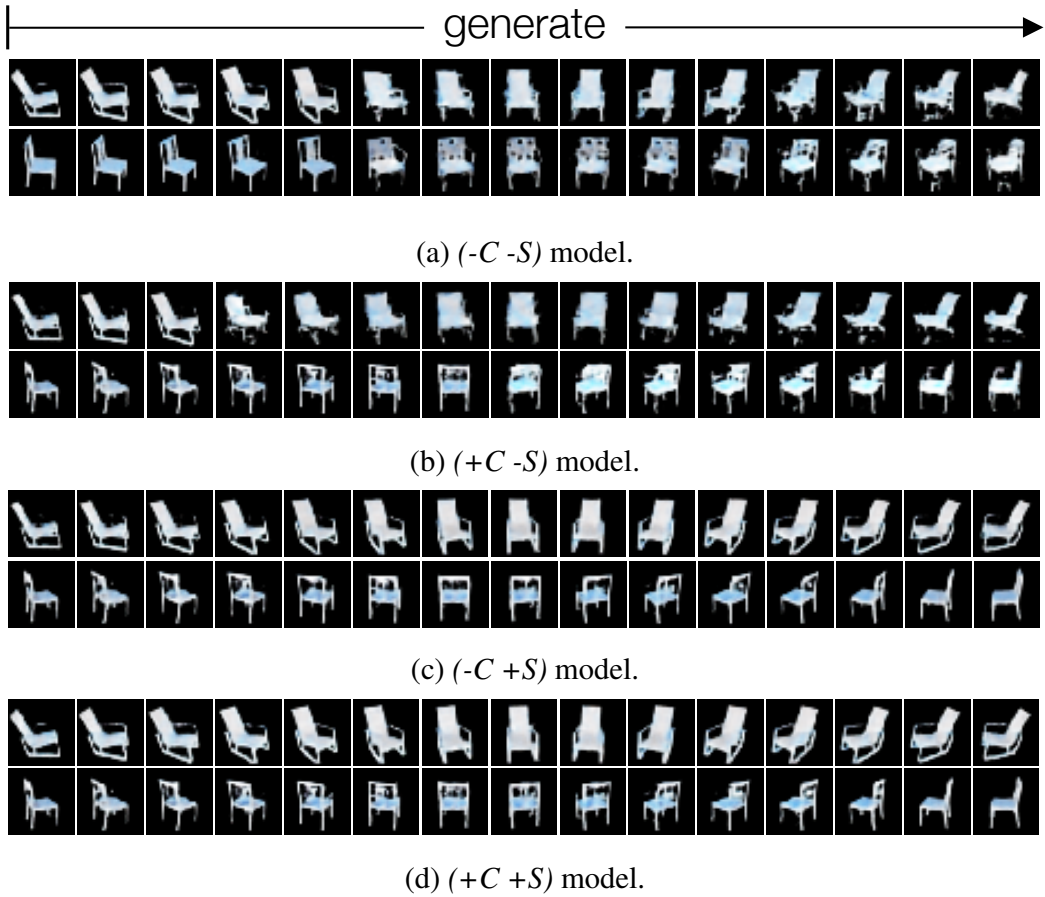

(a) *(-C -S)* model.

(b) *(+C -S)* model.

(c) *(-C +S)* model.

(d) *(+C +S)* model.

Fig. 6: **Visualization of Ablation Study.** We show all four variations of VideoVAE, one per subfigure. Each subfigure contains two synthesized sequences. Observe the substantial improvement due to our two contributions: Employing a structured latent space ($+S$) clearly improves the quality of the generated sequences. Conditional sampling ($+C$) further increases the quality and improves the temporal consistency of the sequences, resulting in our final model, shown in Fig. 6d.

## A.2    Baseline Comparison

The following sections show direct visual comparisons between (**a**) VideoVAE and (**b**) the baselines on all three datasets. As an objective reference, we also include ground truth sequences from the actual datasets, shown in (**c**).

*Chair CAD.*  Qualitative comparison of the generated sequences from our model, Deep Rotator [39], and real videos from Chair CAD [1].

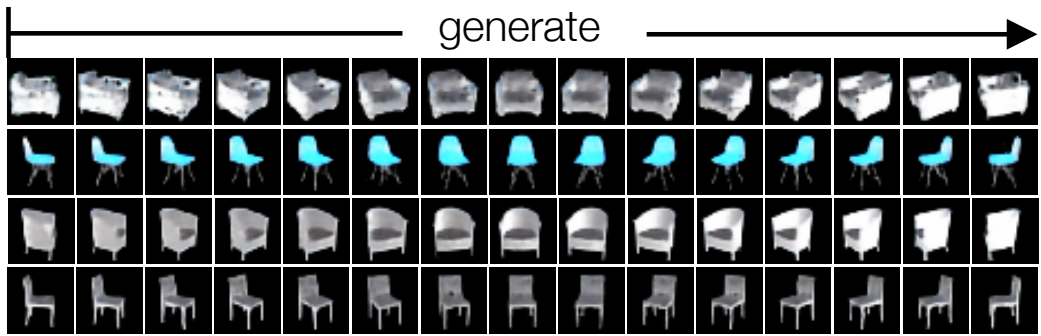

(a) Samples from VideoVAE (**ours**).

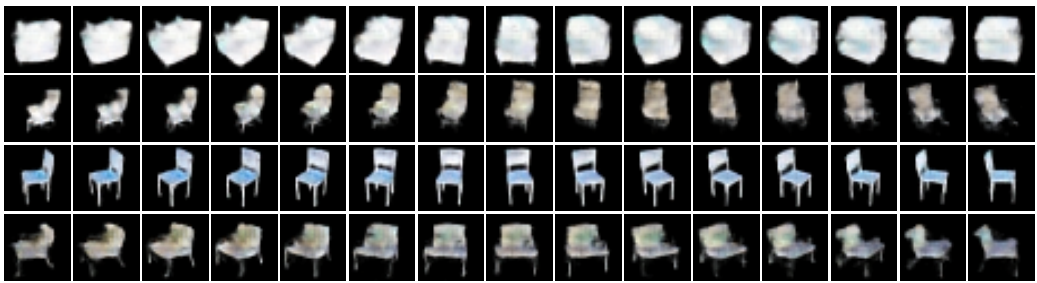

(b) Samples from Deep Rotator [39].

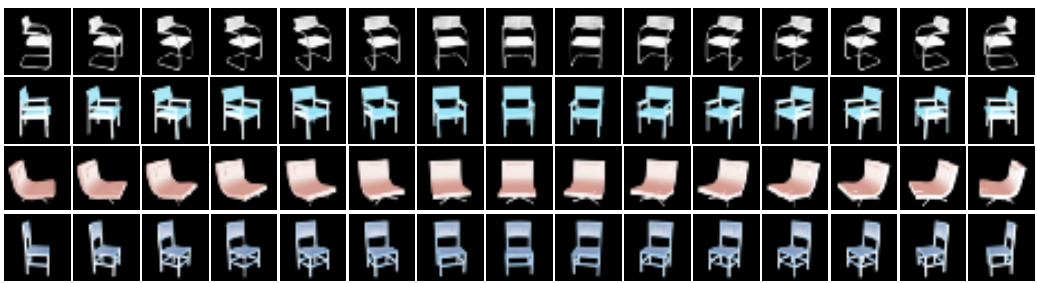

(c) Samples from Chair CAD [1].

Fig. 7: **Qualitative Comparison on Chair CAD.** We show four synthetic sequences from VideoVAE and the Deep Rotator baseline as well as real sequences from Chair CAD. Note the finer details of our model compared to Deep Rotator.

*Weizmann Human Action.* Qualitative comparison of the generated sequences from our model, MoCoGAN [7], and real videos from Weizmann Human Action [2]. Note that the sequences generated by MoCoGAN show mode collapsing behaviour, meaning that all generated sequences fall into a few action classes. Our model's structured latent space manages to preserve all action categories, effectively avoiding mode collapsing. This is also reflected in the *Inter-E* score shown in Table 2 of the main paper (MoCoGAN: 2.91; VideoVAE: 4.37). In addition to the visualizations below, this supplementary material also contains MP4 videos with transient control.[3]

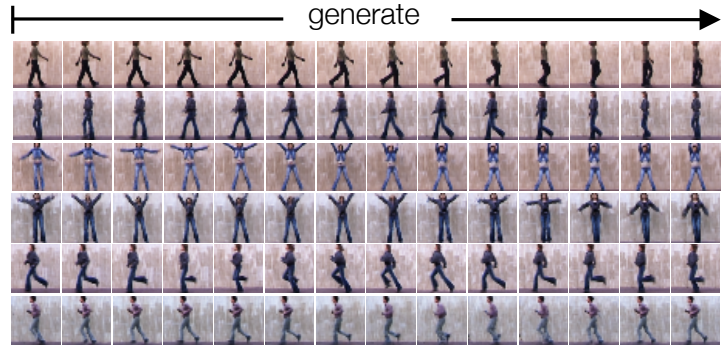

(a) Samples from VideoVAE (**ours**).

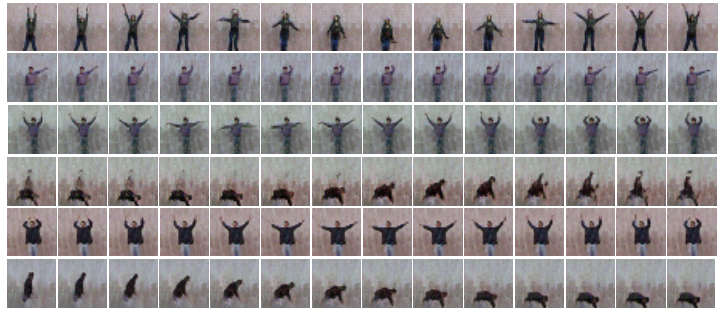

(b) Samples from MoCoGAN [7].

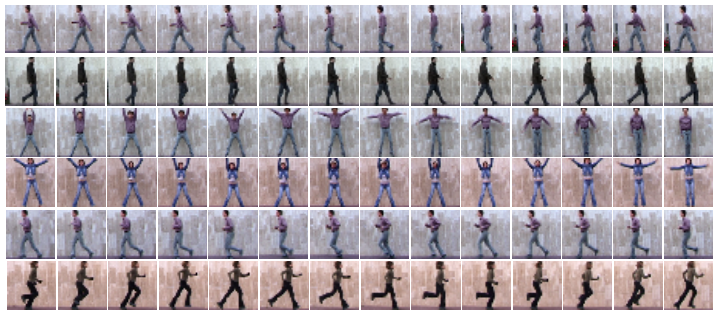

(c) Samples from Weizmann Human Action [2].

Fig. 8: **Qualitative Comparison on Weizmann Human Action.** We show six synthetic sequences from VideoVAE and the MoCoGAN baseline as well as real sequences from Weizmann Human Action.

---

[3] All videos were created using *ffmpeg* with *H.264* codec, which can be downloaded from http://www.divx.com/en/software/technologies/h264

*YFCC-MIT Flickr.* Qualitative comparison of the generated sequences from our model, VGAN [16], and real videos from YFCC-MIT Flickr [3]. Note that although the videos generated by VGAN usually contain noticeable foreground objects and motion, they are less realistic than the ones produced by VideoVAE. Compare Fig. 9a (VideoVAE) and Fig. 9b (VGAN) to the characteristics of actual videos (Fig. 9c).

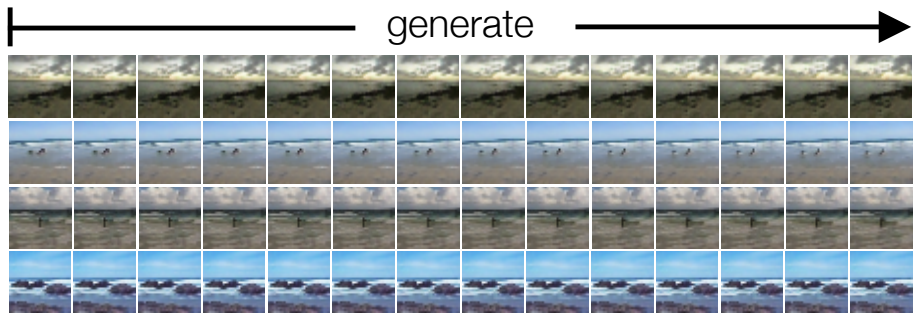

(a) Samples from VideoVAE (**ours**).

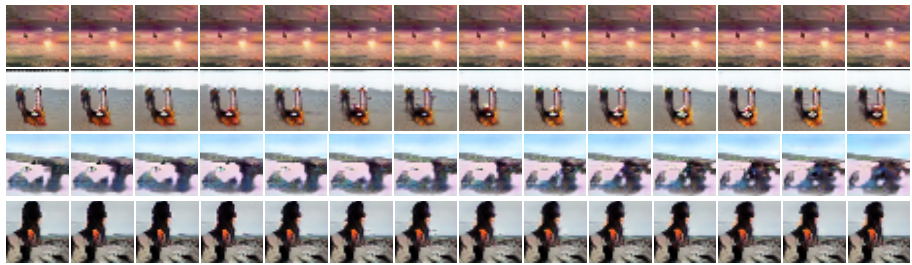

(b) Samples from VGAN [16].

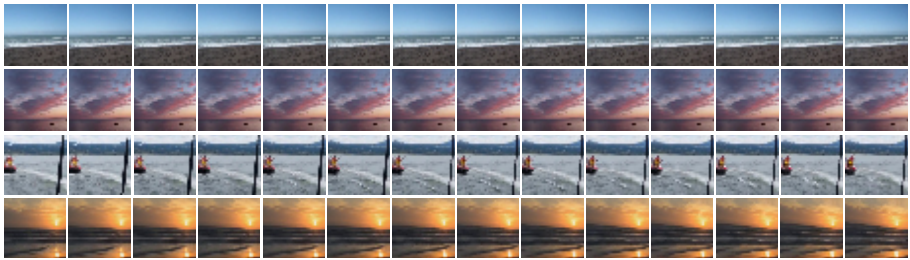

(c) Samples from YFCC-MIT Flickr [3].

Fig. 9: **Qualitative Comparison on YFCC-MIT Flickr.** We show four synthetic sequences from VideoVAE and the VGAN baseline as well as real sequences from YFCC-MIT Flickr.

### A.3    Objective Function Derivation

We derive the variational objective in the dynamic case (Eq. 9). A dynamic latent variable model can be expressed generally as

$$p_\theta(x^{\leq T}, z^{\leq T}) = \prod_{t=1}^{T} p_\theta(x^t, z^t | x^{<t}, z^{<t}) \tag{13}$$

$$= \prod_{t=1}^{T} p_\theta(x^t | x^{<t}, z^{\leq t}) p_\theta(z^t | x^{<t}, z^{<t}). \tag{14}$$

Learning and inference require evaluating the marginal likelihood

$$p_\theta(x^{\leq T}) = \int p_\theta(x^{\leq T}, z^{\leq T}) dz^{\leq T}. \tag{15}$$

To avoid the intractable integration over $z^{\leq T}$, variational inference introduces an approximate posterior, $q_\phi(z^{\leq T} | x^{\leq T})$, which provides a lower bound on $\log p_\theta(x^{\leq T})$:

$$\log p_\theta(x^{\leq T}) \geq \mathcal{L} = \mathbb{E}_{q(z^{\leq T} | x^{\leq T})} \left[ \log \frac{p_\theta(x^{\leq T}, z^{\leq T})}{q_\phi(z^{\leq T} | x^{\leq T})} \right]. \tag{16}$$

In the *filtering* setting, the approximate posterior is only a function of past and present information, and assuming a factorization across time steps yields

$$q_\phi(z^{\leq T} | x^{\leq T}) = \prod_{t=1}^{T} q_\phi(z^t | x^{\leq t}, z^{<t}). \tag{17}$$

Plugging Eqs. 14 and 17 into $\mathcal{L}$ (Eq. 16):

$$\mathcal{L} = \mathbb{E}_{q_\phi(z^{\leq T} | x^{\leq T})} \left[ \log \left( \prod_{t=1}^{T} \frac{p_\theta(x^t | x^{<t}, z^{\leq t}) p_\theta(z^t | x^{<t}, z^{<t})}{q_\phi(z^t | x^{\leq t}, z^{<t})} \right) \right] \tag{18}$$

$$= \mathbb{E}_{q_\phi(z^{\leq T} | x^{\leq T})} \left[ \sum_{t=1}^{T} \log \frac{p_\theta(x^t | x^{<t}, z^{\leq t}) p_\theta(z^t | x^{<t}, z^{<t})}{q_\phi(z^t | x^{\leq t}, z^{<t})} \right]. \tag{19}$$

Rearranging the terms inside the log allows us to separate the conditional log-likelihood from the KL-divergence,

$$\begin{aligned} \mathcal{L} = \mathbb{E}_{q_\phi(z^{\leq T} | x^{\leq T})} [ \sum_{t=1}^{T} (\log p_\theta(x^t | x^{<t}, z^{\leq t}) \\ - \mathrm{KL}(q_\phi(z^t | x^{\leq t}, z^{<t}) || p_\theta(z^t | x^{<t}, z^{<t}))) ]. \end{aligned} \tag{20}$$

Notice that each term in the summation only depends on variables up to time $t$. Therefore, the expectation can be simplified to $q_\phi(z^{\leq t} | x^{\leq t})$, yielding Eq. 9.

### A.4   Details on Architecture and Training

We provide the exact architectural setup of the VAE as well as full training details for each dataset.

**Architecture.**

*Encoder:* conv(kernel 9, channels 10, stride 1) → conv(7, 20, 3) → conv(5, 40, 1) → conv(3, 80, 3) → dropout(0.5) → fc(input 1280, output 1024) → ReLU → fc(1024, 512).

*Structured Latent Space:* The number of dimensions for each distribution in the latent space is set to 512. The small neural networks (black boxes) in Fig. 2b consist of two fully-connected layers with a central ReLU unit.

*Temporal Model:* In this work, the temporal model is set up as a one-layer LSTM with 512 hidden units.

*Decoder:* fc(input 512, output 1024) → ReLU → fc(1024, 1280) → conv(kernel 3, 80, 3) → conv(5, 40, 1) → conv(7, 20, 3) → conv(9, 10, 1)

**Training.** We optimize the objective function using Adam with a learning rate of $1e^{-4}$. The attribute controls used for Chair-CAD are "direction of rotation (rotation)", "tilt" and "chair ID". Note that each chair in Chair-CAD has only 3 sequences for training, which is not sufficient to learn an inference network for "chair ID", so we always provide the first frame in the experiments on this dataset (Table 2). The attribute controls used for Weizmann Human Action are "person ID" and "action class". The attribute control for YFCC-MIT Flicker is "scene label".